\documentclass{ifacconf}

\usepackage{graphicx}      % include this line if your document contains figures
\usepackage{natbib} % required for bibliography
\usepackage{svg}
\usepackage{amsmath,amssymb,mathtools}
\usepackage{bbm}
\usepackage{booktabs}
\usepackage{multirow}
\usepackage{stfloats}
\usepackage{subcaption} % allows individual captions per figure
\usepackage{algorithm}
\usepackage{algpseudocode} % modern replacement for algorithmic
\floatname{algorithm}{Algorithm}
\usepackage[hyphens]{url}
\renewcommand{\thealgorithm}{\arabic{algorithm}}
\usepackage{morefloats} % allows more floats
\usepackage{longtable}
\usepackage{booktabs} % you already use this

\allowdisplaybreaks % Allows automatic page breaks in math environments

\begin{document}
\begin{frontmatter}

\title{Embedding Linear Equality Constraints\\ in Probabilistic Neural Networks\\ for Dynamic Modelling\thanksref{footnoteinfo}} 

%\thanks[footnoteinfo]{Sponsor and financial support acknowledgment
%goes here. Paper titles should be written in uppercase and lowercase
%letters, not all uppercase.}

\thanks[footnoteinfo]{Correspondence to a.del-rio-chanona@imperial.ac.uk}

\author[CE]{Matthew Marsh}
\author[CE]{Beno\^it Chachuat}
\author[CE]{Antonio del Rio Chanona}

\address[CE]{The Sargent Centre for Process Systems Engineering, Department of Chemical Engineering, Imperial College London, UK}

\begin{abstract}                % Abstract of 50--100 words
Machine learning models are increasingly used to model chemical process systems, yet they often lack principled uncertainty quantification and mechanisms to enforce physical constraints. We propose a probabilistic neural network framework that guarantees satisfaction of linear equality constraints within a given tolerance, while capturing aleatoric uncertainty. Compared to state-of-the-art methods, our formulation demonstrates improved predictive accuracy, uncertainty calibration, and adherence to constraints on reduced data. It also demonstrates competitive performance, but with significantly faster training times when evaluated on large data regimes. We evaluated this on two batch reactor case studies, enforcing mass balances.
\end{abstract}


\end{frontmatter}
%===============================================================================

\section{Introduction}

Artificial intelligence (AI) and machine learning (ML) provide powerful methods and tools for learning complex relationships between variables, which are often prevalent in process systems. They have been deployed across various applications for surrogate modelling of processes, including optimisation, monitoring and control \citep{gonzalezNeuralNetworksFast2024, ahmedComparativeStudyMachine2025, lawrenceMachineLearningIndustrial2024}. One major advantage of these approaches is the reduced need for specialist knowledge about an underlying process to build a model around it, as they build directly on data. However, this is also one of their larger drawbacks, namely the lack of integrated physical knowledge. A typical workflow divides the data into `train' and `test' splits, where the model parameters are updated to minimise an error term across the train data and the test data is used to validate the out-of-distribution accuracy of the trained model. Clearly, the performance of a model can be greatly diminished when the volume of data is limited, and its predictions may become inconsistent with known physical laws as a result.

Physics-informed neural networks (PINNs) \citep{raissiPhysicsinformedNeuralNetworks2019} were introduced as a method to improve physical consistency with model predictions. During their training, a term is added to the model loss function that penalises any constraint violations. Empirically, this has shown improved model performance, however these are \textit{soft-constraints}: they do not guarantee constraint satisfaction at training or inference points. Additionally, they increase the number of hyperparameters and may hamper convergence during training. Conversely, \textit{hard-constrained} neural networks enforce constraints at all points. When the constraints define simple lower or upper bounds on the model outputs, these may be enforced directly using a suitable activation function in the output layer. Other methods utilise constraints on model weights to indirectly restrict the output prediction, such as majority vote systems \citep{balestrieroPOLICEProvablyOptimal2023} or positivity to enforce monotonicity or affine constraints \citep{runjeConstrainedMonotonicNeural2023, tordesillasRAYENImpositionHard2023}. 

Alternative approaches augment the model to predict in a reduced subspace of the output variables, with this subset propagated through the constraints to recover the full output space \citep{dontiDC3LearningMethod2021}. However, this may reduce model expressivity and propagate errors in model predictions to the full output. A further method exploits convex optimisation to project the output onto the closest point that satisfies the constraints \citep{agrawalDifferentiableConvexOptimization2019, chenPhysicsinformedNeuralNetworks2024}. For linear equality constraints, this optimisation problem yields a closed-form projection, which does not add to the computational burden during training and inference. Extensions of this method to incorporate inequality or non-linear equality constraints \citep{iftakherPhysicsInformedNeuralNetworks2025, lastrucciENFORCEExactNonlinear2025} have relied either on linear approximations or iterative methods, which can remove constraint satisfaction guarantees and significantly increase the computational burden during both training and inference though.

In practice, the deployment of ML models is often deterministic, which reduces knowledge on the prediction uncertainty. Since the underlying data training these models are often partially observable, noisy or stochastic, deterministic point predictions may be insufficient, particularly when looking to apply these models within an optimisation or control framework.

Uncertainty within data-driven models is commonly attributed to two different, separable sources: aleatoric and epistemic uncertainty. Aleatoric is inherent, irreducible uncertainty, stemming from within the data itself. This is often as a result of noisy measurements, or lack of observability. Epistemic is reducible uncertainty stemming from the model itself, spanning both model architecture choices and parameter values as a result of the training procedures. Both uncertainties can be formulated within a Bayesian inference framework, where the predictive distribution is a function of the likelihood, parametrised by the model itself (i.e., accounting for the aleatoric uncertainty) and a posterior distribution of the parameters over the data itself. Typically, this posterior is intractable for neural networks, and various methods exist for approximating or sampling \citep{gawlikowskiSurveyUncertaintyDeep2022}.

Most approaches to embedding constraints have been enforced deterministically. Some work is also exploring this application in a probabilistic setting, where data-driven models are used to learn solutions to partial differential equations \citep{utkarshEndtoEndProbabilisticFramework2025, hansenLearningPhysicalModels2023}. 

This paper presents a probabilistic neural network framework capable of modelling output aleatoric uncertainty and computing output predictions that are consistent with linear equality constraints across both inputs and outputs, utilizing non-trainable conditioning layers on the output of the model. We explore how these can be used in modelling applications with limited or poor-quality data, and we demonstrate this application within two different batch reactors.

\section{Methodology}

Instead of simple point predictions, probabilistic neural networks can model the aleatoric uncertainty component by serving as a predictor for the parameters or moments of an output data distribution. Neural networks trained on the standard mean-squared error loss function model assume constant variance in the output. Alternatively, we assume here that the covariance of the output distribution is learnable, and we try to predict it via a second output head alongside the output mean ($\boldsymbol{\mu}_P$). As the covariance matrix ($\boldsymbol{\Sigma}_P$) must be symmetric and positive semi-definite, we enforce this second output head to predict the Cholesky factor of the covariance matrix ($\mathbf{L}_P$). Both output heads are then used to minimise the negative log-likelihood (NLL) of a multivariate Gaussian distribution as the loss function, with the neural network $\mathbf{f}_{\boldsymbol{\theta}}$ parametrising this unconstrained distribution:
\begin{equation}
    \label{eq:unconstrained}
    p(\mathbf{y} \mid \mathbf{x}) \coloneqq \mathcal{N}(\boldsymbol{\mu}_P, \mathbf{L}_P\mathbf{L}_P^\intercal) \quad \text{with}~~(\boldsymbol{\mu}_P, \mathbf{L}_P) = \mathbf{f}_{\boldsymbol{\theta}}(\mathbf{x})
\end{equation}
for given inputs $\mathbf{x} \in \mathbb{R}^{n_x}$ and learnable outputs $\mathbf{y} \in \mathbb{R}^{n_y}$. 

Given a set of independent linear equality constraints (assuming $\mathbf{B}$ is full-rank)
\begin{equation}
    \mathbf{A}\mathbf{x} + \mathbf{B}\mathbf{y} = \mathbf{b},
\end{equation}
we define the residual of the constraint as $\mathbf{z}\in\mathbb{R}^{n_c}$ with $n_c<n_y$, which is described by a delta distribution when both $\mathbf{x}$ and $\mathbf{y}$ are realised. To recast the distribution of $\mathbf{z}$ as a Gaussian and allow for tractable, closed-form conditioning, we introduce a small numerical diffusion term $\boldsymbol{\epsilon}$, that defines a `tolerance' around the constraint. In contrast to \citep{marshLearningEmbeddedLinear2026a}, we treat this term as a hyperparameter. 
\begin{equation}
    \mathbf{z} = \mathbf{A}\mathbf{x} + \mathbf{B}\mathbf{y} - \mathbf{b} + \boldsymbol{\epsilon} \quad 
    \text{with}~~~\boldsymbol{\epsilon} \sim \mathcal{N}(\mathbf{0}, \epsilon \mathbf{I}).
\end{equation}
%As the models are trained on noisy data, this also prevents noise being propagated through the constraints.

Since $\mathbf{z}$ is now normally distributed,
\begin{equation}
    p(\mathbf{z} \mid \mathbf{y}, \mathbf{x}) = \mathcal{N}(\mathbf{A}\mathbf{x} + \mathbf{B}\mathbf{y} - \mathbf{b}, \epsilon \mathbf{I}),
\end{equation}
we can exploit known results from linear Gaussian systems to form a joint distribution between the residual variable and the neural network output, 
\begin{equation}
p(\mathbf{y}, \mathbf{z} \!\mid\! \mathbf{x}) = \mathcal{N} \!\left(\!
\begin{bmatrix}
\boldsymbol{\mu}_P \\
\mathbf{A} \mathbf{x} + \mathbf{B} \boldsymbol{\mu}_P - \mathbf{b}
\end{bmatrix}\!,\!
\begin{bmatrix}
\boldsymbol{\Sigma}_P & \boldsymbol{\Sigma}_P \mathbf{B}^\intercal \\
\mathbf{B} \boldsymbol{\Sigma}_P & \mathbf{B} \boldsymbol{\Sigma}_P \mathbf{B}^\intercal + \epsilon \mathbf{I}
\end{bmatrix}
\!\right)
\end{equation}
for a specified input value $\mathbf{x}$. 

Enforcement of the equality constraint (within the tolerance region) entails conditioning the unconstrained output distribution~\eqref{eq:unconstrained} on $\mathbf{z}=0$,
\begin{equation}
    \label{eq:conditioned}
    p(\mathbf{y} \mid \mathbf{z=0}, \mathbf{x}) \eqqcolon \mathcal{N}(\boldsymbol{\mu}_Q, \boldsymbol{\Sigma}_Q),
\end{equation}
which is again a Gaussian distribution, with the following closed-form updates for the mean and covariance matrix:
\begin{align*}
    \boldsymbol{\mu}_Q & = \boldsymbol{\mu}_P + \boldsymbol{\Sigma}_P \mathbf{B}^\intercal 
    \left(\mathbf{B}\boldsymbol{\Sigma}_P \mathbf{B}^\intercal + \epsilon \mathbf{I}\right)^{-1} 
    \left(\mathbf{b} - \mathbf{A}\mathbf{x} - \mathbf{B}\boldsymbol{\mu}_P \right)\\
    \boldsymbol{\Sigma}_Q & = \boldsymbol{\Sigma}_P - \boldsymbol{\Sigma}_P \mathbf{B}^\intercal 
    \left(\mathbf{B}\boldsymbol{\Sigma}_P \mathbf{B}^\intercal + \epsilon \mathbf{I}\right)^{-1} 
    \mathbf{B}\boldsymbol{\Sigma}_P.
\end{align*}

The output predictor is then given by the conditional distribution \eqref{eq:conditioned}. Conditioning $p(\mathbf{y}, \mathbf{z} \mid \mathbf{x})$ on the constraint is equivalent to projecting the joint onto the constraint manifold using the KL divergence and yields the above updates for linear–Gaussian models. This projects the mean onto the manifold where the constraint is satisfied and reduces the uncertainty in the direction of the constraint to the tolerance $\epsilon$.  A quick sanity check confirms that
\[ \mathbf{A}\mathbf{x}+\mathbf{B}\boldsymbol{\mu}_Q \to \mathbf{b} \qquad \text{and} \qquad \mathbf{B}\boldsymbol{\Sigma}_Q\mathbf{B}^\intercal \to \mathbf{0} \]
as the tolerance $\epsilon\to 0$.

Therefore, the resulting \textit{constrained probabilistic neural network} (CPNN) can be trained on any data set $\mathcal{D} \coloneqq \{(\mathbf{x}_i, \mathbf{y}_i)\}_{i=1}^{n_d}$ by minimising the conditioned Gaussian NLL loss, given by
%\begin{equation}
%    \mathcal{L}\!\left(\mathbf{y} \mid \boldsymbol{\mu}_Q, \boldsymbol{\Sigma}_Q\right) = \tfrac{1}{2} (\mathbf{y} - {\boldsymbol{\mu}}_Q)^\intercal 
%    {\boldsymbol{\Sigma}}_Q^{-1} (\mathbf{y} - {\boldsymbol{\mu}}_Q) 
%    + \tfrac{1}{2} \log \det {\boldsymbol{\Sigma}}_Q.
%\end{equation}
\begin{align}
    \mathcal{L}\!\left(\mathcal{D} \mid \boldsymbol{\mu}_Q, \boldsymbol{\Sigma}_Q\right) =\:& \tfrac{1}{2}\!\!\!\!\!\! \sum_{(\mathbf{x}_i, \mathbf{y}_i)\in\mathcal{D}}\!\!\!\!\! (\mathbf{y}_i - {\boldsymbol{\mu}}_Q)^\intercal 
    {\boldsymbol{\Sigma}}_Q^{-1} (\mathbf{y}_i - {\boldsymbol{\mu}}_Q)\nonumber\\ &
    + \tfrac{1}{2} \log \det {\boldsymbol{\Sigma}}_Q. \label{eq:NLLloss}
\end{align}
where both $\boldsymbol{\mu}_Q$ and $\boldsymbol{\Sigma}_Q$ depend on the input $\mathbf{x}$. Algorithm~\ref{alg:training} summarises the overall training procedure.
\bigskip

\begin{algorithm}[htb]
\caption{Training Procedure for CPNN}\small
\label{alg:training}
\begin{algorithmic}[1]
\Require Training data $\mathcal{D} = \{(\mathbf{x}_i, \mathbf{y}_i)\}_{i=1}^{n_d}$, constraint matrices $(\mathbf{A}, \mathbf{B}, \mathbf{b})$, diffusion parameter $\epsilon$
\State Initialize model $\mathbf{f}_{\boldsymbol{\theta}}$ with feedforward backbone and conditioning layers
\State Set learning rate $\eta$, total epochs $n_e$, and optimizer $\mathcal{O}$

\For{$e = 1$ \textbf{to} $n_e$}
    \State Sample mini-batch $\{(\mathbf{x}_b, \mathbf{y}_b)\} \eqqcolon \mathcal{D}_b \subset \mathcal{D}$
    \State Compute unconstrained outputs $(\boldsymbol{\mu}_P, \mathbf{L}_P) = \mathbf{f}_{\boldsymbol{\theta}}(\mathbf{x}_b), \forall b$
    \State Apply conditioning:% step and compute overall output:
    \[
    \begin{aligned}
        \boldsymbol{\mu}_Q &= \boldsymbol{\mu}_P 
        + \boldsymbol{\Sigma}_P \mathbf{B}^\intercal 
        \left(\mathbf{B}\boldsymbol{\Sigma}_P \mathbf{B}^\intercal + \epsilon \mathbf{I}\right)^{-1}
        \left(\mathbf{b} - \mathbf{A}\mathbf{x}_b - \mathbf{B}\boldsymbol{\mu}_P \right) \\
        \boldsymbol{\Sigma}_Q &= \boldsymbol{\Sigma}_P 
        - \boldsymbol{\Sigma}_P \mathbf{B}^\intercal 
        \left(\mathbf{B}\boldsymbol{\Sigma}_P \mathbf{B}^\intercal + \epsilon \mathbf{I}\right)^{-1}
        \mathbf{B}\boldsymbol{\Sigma}_P
    \end{aligned}
    \]
    \State Compute:\ $\mathcal{L}\!\left(\mathcal{D}_b \mid \boldsymbol{\mu}_Q, \boldsymbol{\Sigma}_Q\right)$,\ $\boldsymbol{\nabla}_{\boldsymbol{\theta}} \mathcal{L}\!\left(\mathcal{D}_b \mid \boldsymbol{\mu}_Q, \boldsymbol{\Sigma}_Q\right)$
    \State Update parameters:\ $\boldsymbol{\theta} \leftarrow \mathcal{O}\!\left(\boldsymbol{\theta}, \boldsymbol{\nabla}_{\boldsymbol{\theta}} \mathcal{L}, \eta\right)$
\EndFor

\State {\bf Return:} Trained model $\mathbf{f}_{\boldsymbol{\theta}}$
\end{algorithmic}
\end{algorithm}

\textbf{Implementation. } All models were implemented and trained in \texttt{Python} using \texttt{PyTorch} \citep{pytorch_2019}. The conditioning procedure was realised through two custom differentiable layers, designed to be non-trainable yet fully embedded within the model’s computational graph. During each forward pass, the unprojected mean vector and Cholesky factor of the covariance matrix are provided as inputs to these layers. The conditioning layers are defined \textit{a priori} to model training, initialised with the known constraint matrices and chosen tolerance regions ($\boldsymbol{\epsilon}$). This design ensures that the projection step remains both tractable and differentiable throughout training. Each model employed a feedforward artificial neural network (ANN) backbone with ReLU activation functions between layers, each trained on 150 epochs. The architectural hyperparameters---including the number of hidden layers, the number of hidden units, and the learning parameters---were tuned using the validation loss in \texttt{Optuna} \citep{optuna_2019}. All training was carried out on a 2023 MacBook Pro with M2 Pro chip and 16GB RAM.

\section{Case Studies}

We evaluate the framework on two batch reactor case studies. Batch reactors are common in fine chemicals and pharmaceuticals, but their transient dynamics, strong temperature–concentration coupling, and lack of steady-state operation make them challenging to model and control when the underlying dynamics are not well understood. These properties make them suitable benchmarks for data-driven modelling under uncertainty.

We used a full mechanistic dynamic model to generate the datasets to train and validate the models on. As both case studies are batch systems, the full datasets were made up of multiple batch cycles. The first case study utilised a cycle of 300 time steps and the second of 100, both initialised using random initial conditions and perturbed with random action step changes throughout each simulation, with added Gaussian noise of 5\%. Each model was trained on four different data regimes, to investigate how this affected their performance. The {\sf minimal} regime utilised one simulation each for training, test and validation. The other data regimes split the first 60\% of the simulations for training, 20\% for hyperparameter and model tuning, and the final 20\% for testing, with the {\sf small} data regime consisting of 10 overall simulations, 50 for the {\sf medium} regime, and 100 for the {\sf large} regime. 

\subsection{Non-Isothermal Batch Reactor with Irreversible Reactions}

The first system considers a jacketed batch reactor, with two consecutive, irreversible, exothermic reactions:
\begin{center}
$A \rightarrow B \rightarrow C$
\end{center}
The dynamic model consists of four state variables; the concentration of each species in the reactor ($c_i, i=A,B,C)$ and the temperature ($T$); and one control variable, the cooling jacket temperature ($T_c$).
The nonlinear dynamics follow standard material and energy balances with first-order kinetic rates:
\begin{align}
\label{eq:dCA_sub}
\tfrac{dc_A}{dt} & =
-k^0_{1} \exp\!\left(-\tfrac{E_{1}}{RT}\right) c_A\\
\label{eq:dCB_sub}
\tfrac{dc_B}{dt} & = 
2k^0_{1} \exp\!\left(-\tfrac{E_{1}}{RT}\right) c_A 
- k^0_{2} \exp\!\left(-\tfrac{E_{2}}{RT}\right) c_B\\
\label{eq:dCC_sub}
\tfrac{dc_C}{dt} & = 
k^0_{2} \exp\!\left(-\tfrac{E_{2}}{RT}\right) c_B\\
\label{eq:dT_sub_compact}
\tfrac{dT}{dt} &= \sum_{i=1,2} \tfrac{-\Delta H_i}{\rho C_p} k^0_{i} \exp\!\left(-\tfrac{E_{i}}{RT}\right) c_i + \tfrac{UA}{\rho C_p V}(T_c - T)
\end{align}
Kinetic constraints are set to $k^0_1=1~\text{s}^{-1}$, $k^0_2=0.5~\text{s}^{-1}$, $E_1=5000~\text{kJ\,kmol}^{-1}$, and $E_2=6000~\text{kJ\,kmol}^{-1}$. The initial concentrations were bounded between 0 and 1, with the initial temperatures and action spaces bounded between $250\,^\circ\mathrm{C}$ and $450\,^\circ\mathrm{C}$. The reaction system conserves the following linear invariant:
\[ h(\mathbf{c}) \coloneqq 2c_A+c_B+c_C. \]
The CPNN model takes the current state and control as an input, and aims to predict the next state:
\begin{equation*} \label{eq:state_output_vectors}
\begin{aligned}
\mathbf{x} &= [\,c_{A,t},\, c_{B,t},\, c_{C,t},\, T_t,\, T_{c,t}\,], \\
\mathbf{y} &= [\,c_{A,t+1},\, c_{B,t+1},\, c_{C,t+1},\, T_{t+1}\,],
\end{aligned}
\end{equation*}
while enforcing a mass balance across time steps as an affine constraint:
\[
\mathbf{A} = \begin{bmatrix} -2 & -1 & -1 & 0 & 0 \end{bmatrix}, \quad
\mathbf{B} = \begin{bmatrix} 2 & 1 & 1 & 0 \end{bmatrix}, \quad
\mathbf{b} = \begin{bmatrix} 0 \end{bmatrix}.
\]

\subsection{Isothermal Batch Reactor with Reversible Reactions}

The second reactor is isothermal, containing four states and no control inputs. The species interact via two reversible reactions \citep{villanuevaStabilitySetValuedIntegration2014}:

\begin{center}
$A + B \rightleftharpoons C, \qquad A + C \rightleftharpoons D$
\end{center}

The nonlinear dynamics of the system are described by:
\begin{align}
\tfrac{dx_A}{dt} &= - (k_{1}^f x_A x_B - k_{1}^r x_C) - (k_{2}^f x_A x_C - k_{2}^r x_D) \\
\tfrac{dx_B}{dt} &= - (k_{1}^f x_A x_B - k_{1}^r x_C) \\
\tfrac{dx_C}{dt} &= (k_{1}^f x_A x_B - k_{1}^r x_C) - (k_{2}^f x_A x_C - k_{2}^r x_D) \\
\tfrac{dx_D}{dt} &= k_{2}^f x_A x_C - k_{2}^r x_D
\end{align}
where $x_i$ are mole fractions for each species, $i=A,B,C,D$; and kinetic constants are set to $k_2^f = 2$, $k_1^r = k_2^r = 1$, with $k_1^f \sim \mathcal{U}[50,60]$, which acts as an additional aleatoric uncertainty. Due to the reversible reactions, the system conserves two linear reaction invariants:
\begin{align*}
    h_1(\mathbf{x}) &= x_B + x_C + x_D, \quad h_2(\mathbf{x}) = x_A - x_B + x_D,
\end{align*}
which are embedded as constraints within the CPNN:
\begin{align*}
\mathbf{A} &=
\begin{bmatrix}
0 & -1 & -1 & -1 \\
-1 & 1 & 0 & -1
\end{bmatrix},
\quad
\mathbf{B} =
\begin{bmatrix}
0 & 1 & 1 & 1 \\
1 & -1 & 0 & 1
\end{bmatrix}, 
\quad
\mathbf{b} = \begin{bmatrix}
0 \\
0 
\end{bmatrix}.
\end{align*}
The model setup again considers one-step ahead predictions:
\begin{align*} %\label{eq:state_constraints}
\mathbf{x} &= [x_{A,t},\, x_{B,t},\, x_{C,t},\, x_{D,t}],\\
\mathbf{y} &= [x_{A,t+1},\, x_{B,t+1},\, x_{C,t+1},\, x_{D,t+1}].
\end{align*}

\section{Results}

\begin{figure*}[tb] 
    \centering
    \includegraphics[width=0.46\linewidth]{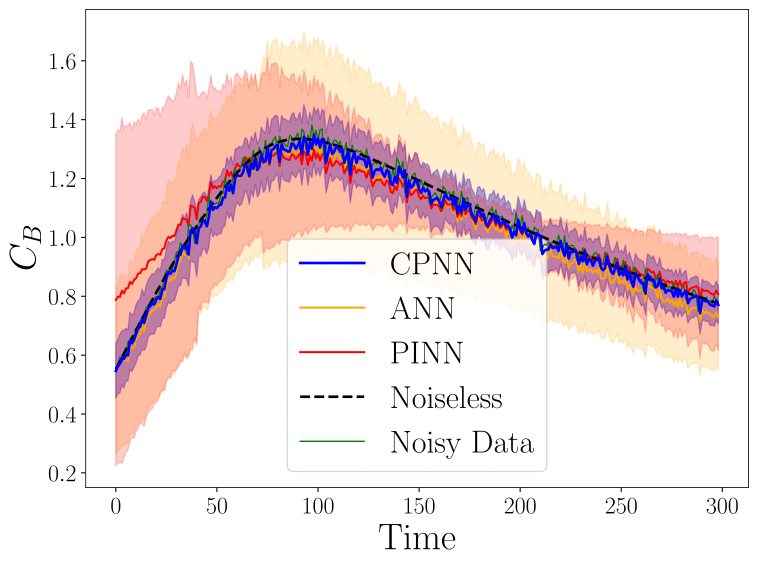}
    \hfill
    \includegraphics[width=0.46\linewidth]{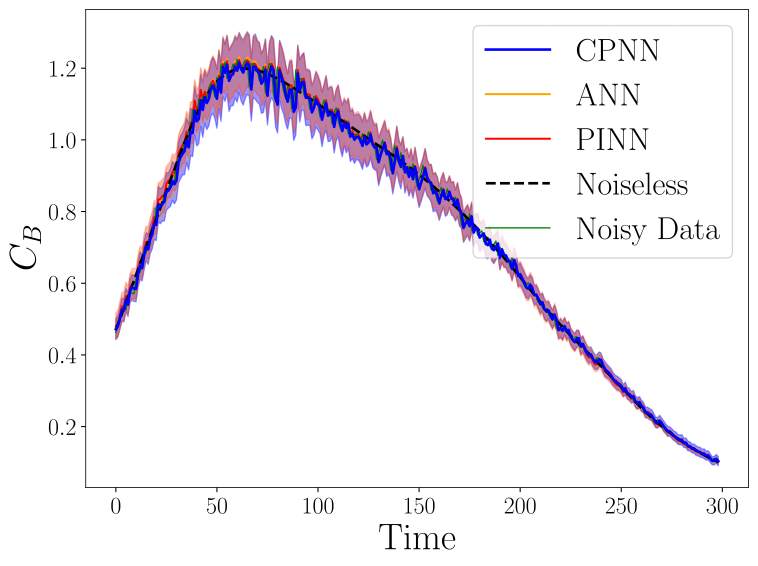}
    \vspace{-.5em}
    \caption{$c_B$ predictions on {\sf minimal} (left) and {\sf large} (right) data regimes for the batch reactor with irreversible reactions.}
    \label{fig:CB}
\end{figure*}

We benchmarked the CPNN against a standard feedforward ANN and a soft-constrained PINN, where the loss function was penalised with the residual of the constraints, using the mean. All models were trained on the multivariate Gaussian NLL loss function \eqref{eq:NLLloss}.

We split our evaluation into four distinct categories. Firstly, we assess how accurate the predicted mean is from the underlying \textit{noiseless data}, using mean squared error (MSE). Secondly, we assess the quality of the \textit{predictive distribution} against the underlying data, using the continuous ranked probability score (CRPS) compared to the \textit{noiseless data}; the coverage ratio and width are also compared to the \textit{noisy data}, testing whether 95\% of the data fall within the theoretical 95\% bounds. Thirdly, we compare the mean constraint violation across the model predictions on the test set. Finally, we monitor the time it took to train each of the models across each regime. All the results are summarised in Tables~\ref{tab:complexity} \& \ref{tab:metrics-merged}.

Across both case studies, the dominant factor affecting model performance was the data regime. When training on the {\sf minimal} data regime, the CPNN strongly outperforms the baseline models (ANN, PINN), as they are unable to generalise to unseen data from a single training simulation. When data is increased to the {\sf small} data regime, the CPNN still consistently outperforms the baselines across all evaluation metrics, albeit to a lesser degree. This advantage stems from embedding the constraints directly into the learning process, which reduces overfitting and guides the model towards physically feasible solutions. The effect is more pronounced when enforcing more constraints, as more information is injected into the model via the projection step.

When data is further increased to the {\sf medium} regime, the unconstrained models begin to learn the invariants, narrowing the performance gap, reducing the advantage gained when explicitly embedding physical knowledge. The ANN and soft-constrained PINN approach the accuracy of the CPNN with regards to point prediction and learning the constraints between the variables. In the {\sf large} data regime, the unconstrained models achieved similar or slightly better accuracy than the CPNN, as sufficient data allowed them to learn relationships without explicit embedding.

\begin{figure*}[tb] 
    \centering
    \includegraphics[width=0.46\linewidth]{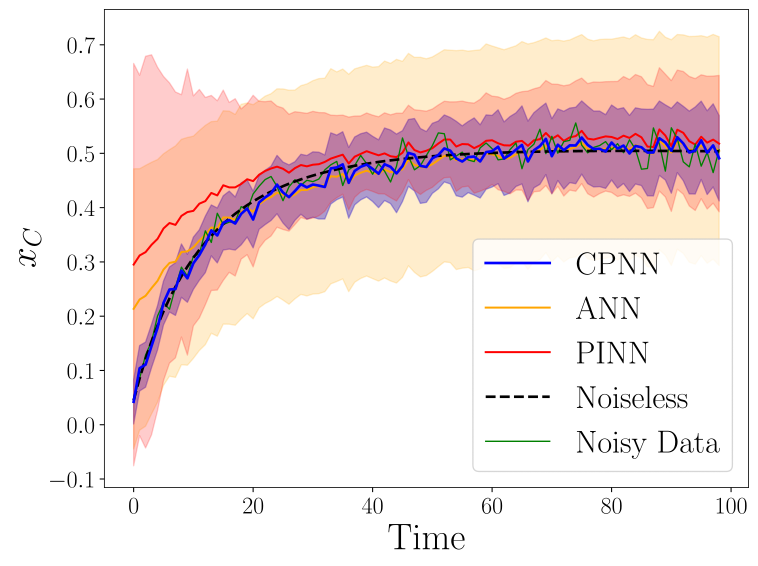}
    \hfill
    \includegraphics[width=0.46\linewidth]{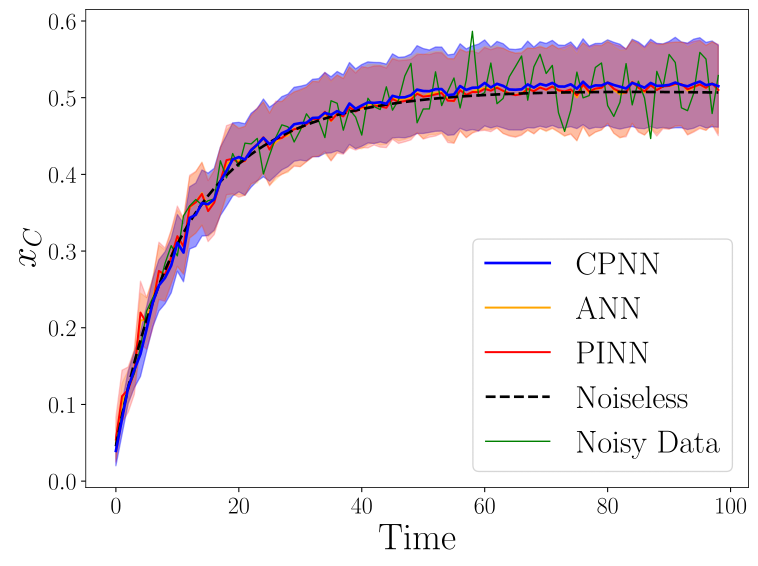}
    \vspace{-.5em}
    \caption{$x_C$ predictions on {\sf minimal} (left) and {\sf large} (right) data regimes for the batch reactor with reversible reactions.}
    \label{fig:xC}
\end{figure*}

Despite the MSE and constraint violation being consistently low, when examining the trajectory prediction plots of selected variables and regimes (see Figs.~\ref{fig:CB} \& \ref{fig:xC}), allows for further analysis. These plots were generated by looping through a simulation in the test set, with the model predicting one step ahead from the current state and appended to build a full trajectory distribution. The shaded region denotes the 95\% confidence region ($\mu \pm 1.96 \sigma$) of the model predictions. The plots show the mean predictions often aligns with the noisy data, rather the true system, which leads to noisy and overestimated confidence intervals, hence poor uncertainty calibration. As the model conditions only on the previous noisy state, rather than an input sequence, it may propagate noise rather than infer the smooth underlying trend. The plots also emphasise the improvements with more data, as across the {\sf minimal} regime, the CPNN is able to predict closer to the true value with a tighter confidence interval than the others, whereas as we move towards the {\sf large} regime, the three model predictions converge, showing reduced advantage to embedding the constraints.

When evaluating the predictive \textit{distribution} in terms of CRPS, similar trends are observed to other metrics. When analysing the coverage width and ratios, the 95\% coverage ratio across all case studies and models is consistently above 0.95. This suggests that the learned variance is over-estimated, leading to poorly calibrated uncertainty intervals and reinforcing the suspicion that the models have a tendency to overfit to the noise of the data rather than learn the true underlying dynamics. As the gradients of the loss functions are dependent on the learned variance, this overfit is difficult to correct. Two methods have been proposed to combat this phenomenon, to reduce the dependency. Firstly, the $\beta$ NLL loss \citep{seitzerPitfallsHeteroscedasticUncertainty2022}, which in effect adds annealing to the learned variance. Other methods \citep{immerEffectiveBayesianHeteroscedastic2023} have investigated reparametrizing the Gaussian using the information or canonical parameters.

\begin{table}[tb]
\centering
\caption{Training times (s) across data regimes.}
\label{tab:complexity}
\setlength{\tabcolsep}{10pt}
\begin{tabular}{lcccc}
\toprule
\textbf{Model} & \textbf{Minimal} & \textbf{Small} & \textbf{Medium} & \textbf{Large} \\
\toprule
\multicolumn{5}{l}{\textbf{Batch Reactor with Irreversible Reactions}}\\
\midrule
CPNN & \textbf{4.41} & \textbf{16.17} & \textbf{71.67} & \textbf{150.10} \\
ANN  & 8.70 & 38.65 & 199.89 & 389.06 \\
PINN & 8.33 & 39.04 & 195.29 & 385.79 \\
\midrule
\multicolumn{5}{l}{\textbf{Batch Reactor with Reversible Reactions}}\\
\midrule
CPNN & \textbf{11.79} & \textbf{45.13} & \textbf{221.03} & \textbf{438.18} \\
ANN  & 27.07 & 121.45 & 531.98 & 1088.69 \\
PINN & 25.40 & 113.48 & 571 & 1128.98 \\
\bottomrule
\end{tabular}
\end{table}

The comparison of computational complexity in Table \ref{tab:complexity}, across each model also shows interesting trends. Although none of these models are overtly complex or utilise huge amounts of data, there are clear differences in the time it took to train each model. The CPNN consistently cuts the training time by half or more on all case studies and regimes, over the same number of epochs. This suggests the conditioning step alters the loss landscape of the CPNN compared to the ANN and PINN, allowing for faster convergence. This could be due to the conditioning step focusing the search towards feasible, and hence better solutions, that result in a lower loss. The embedded constraints also allow for smaller NN architectures, which again reduces computational overhead.

\begin{table*}[tb]   % bottom of page 5
\centering
\caption{Normalised test metrics across all regimes and case studies.}
\label{tab:metrics-merged}
\renewcommand{\arraystretch}{1.05}
\setlength{\tabcolsep}{7pt}

%\begin{tabular*}{\textwidth}{@{\extracolsep{\fill}} l cccc }
\begin{tabular*}{\textwidth}{llr@{$~\pm~$}lr@{$~\pm~$}lr@{$~\pm~$}lr@{$~\pm~$}l }
\toprule
\textbf{Method} & \textbf{Metric} & \multicolumn{2}{c}{\textbf{Minimal}} & \multicolumn{2}{c}{\textbf{Small}} & \multicolumn{2}{c}{\textbf{Medium}} & \multicolumn{2}{c}{\textbf{Large}} \\
\toprule
\multicolumn{10}{l}{\textbf{Non-Isothermal Batch Reactor with Irreversible Reactions}}\\
\midrule

% ======= CPNN =======
\textbf{CPNN} & Mean violation & \textbf{0.0342}  & \textbf{0.0149}   & \textbf{0.0175}  & \textbf{0.0144} & 0.0167 & 0.0105 & 0.0149 & 0.011 \\
& MSE                     & \textbf{0.000357} & \textbf{0.00053} & \textbf{0.000244}  & \textbf{0.000565} & \textbf{0.000114}  & \textbf{0.000229} & 0.000119 & 0.000273 \\
& Coverage Width          & \textbf{0.0787}   & \textbf{0.0407}  & \textbf{0.0841}  & \textbf{0.0547} & \textbf{0.0754}  & \textbf{0.0402} & 0.0812 & 0.048 \\
& Coverage Ratio          & \textbf{0.96}     & \textbf{0.197}   & 0.996 & 0.0605 & 0.999 & 0.010 & 0.998 & 0.0357 \\
& CRPS                    & \textbf{0.0105}   & \textbf{0.00768} & \textbf{0.00817}  & \textbf{0.0065} & \textbf{0.00637}  & \textbf{0.00421} & 0.00662 & 0.00448 \\
\midrule
% ======= MLP =======
\textbf{ANN} & Mean violation & 0.121 & 0.0126 & 0.0296 & 0.0099 & \textbf{0.0143}  & \textbf{0.0104} & \textbf{0.0144}  & \textbf{0.0103} \\
& MSE                 & 0.0107 & 0.0109 & 0.000344 & 0.000538 & 0.000131 & 0.000343 & \textbf{0.000119}  & \textbf{0.000289} \\
& Coverage Width      & 1.18 & 0.79 & 0.135 & 0.0364 & 0.0796 & 0.0483 & \textbf{0.0767}  & \textbf{0.0473} \\
& Coverage Ratio      & 1.00 & 0 & 0.994 & 0.0788 & \textbf{0.998}  & \textbf{0.0334} & \textbf{0.997}  & \textbf{0.0446} \\
& CRPS                & 0.0806 & 0.0526 & 0.0116 & 0.00583 & 0.00661 & 0.0049 & \textbf{0.0064}  & \textbf{0.00466} \\
\midrule
% ======= PINN =======
\textbf{PINN} & Mean violation & 0.0387 & 0.0149 & 0.028 & 0.0164 & 0.0145 & 0.0106 & 0.0149 & 0.0107 \\
& MSE                 & 0.000942 & 0.00249 & 0.000291 & 0.000469 & 0.000123 & 0.000319 & 0.000136 & 0.000349 \\
& Coverage Width      & 0.136 & 0.0656 & 0.0956 & 0.0365 & 0.0816 & 0.049 & 0.0777 & 0.0489 \\
& Coverage Ratio      & 0.997 & 0.0502 & \textbf{0.992}  & \textbf{0.0854} & 0.999 & 0.0151 & 0.997 & 0.0393 \\
& CRPS                & 0.0154 & 0.0124 & 0.00969 & 0.00577 & 0.00662 & 0.00471 & 0.00661 & 0.00502 \\
\toprule
\multicolumn{10}{l}{\textbf{Isothermal Batch Reactor with Reversible Reactions}}\\
\midrule
% ======= CPNN =======
\textbf{CPNN} & Mean violation & \textbf{0.0474} & \textbf{0.0131} & \textbf{0.0285}  & \textbf{0.0189} & \textbf{0.0164}  & \textbf{0.0134} & 0.0196 & 0.0169 \\
& MSE                 & \textbf{0.0016} & \textbf{0.00306} & \textbf{0.000252}  & \textbf{0.000413} & 0.000164 & 0.000332 & 0.000159 & 0.000326 \\
& Coverage Width      & \textbf{0.124} & \textbf{0.0594} & 0.0851 & 0.0555 & 0.0624 & 0.0402 & 0.0611 & 0.0386 \\
& Coverage Ratio      & \textbf{0.966} & \textbf{0.182} & \textbf{0.949}  & \textbf{0.195} & 0.993 & 0.0749 & 0.993 & 0.0792 \\
& CRPS                & \textbf{0.0195} & \textbf{0.0196} & \textbf{0.00902}  & \textbf{0.00645} & 0.00644 & 0.00571 & 0.00633 & 0.00564 \\
\midrule
% ======= MLP =======
\textbf{ANN} & Mean violation & 0.127 & 0.0695 & 0.0453 & 0.019 & 0.0166 & 0.0136 & 0.0167 & 0.0135 \\
& MSE                 & 0.014 & 0.0223 & 0.000624 & 0.001 & \textbf{0.000161}  & \textbf{0.000329} & 0.000146 & 0.000295 \\
& Coverage Width      & 0.59 & 0.444 & 0.147 & 0.0659 & \textbf{0.0624}  & \textbf{0.0417} & \textbf{0.0605}  & \textbf{0.0405} \\
& Coverage Ratio      & 0.723 & 0.447 & 0.997 & 0.0408 & 0.893 & 0.300 & 0.96 & 0.17 \\
& CRPS                & 0.0745 & 0.0546 & 0.0142 & 0.00874 & \textbf{0.00634}  & \textbf{0.00568} & 0.00613 & 0.00541 \\
\midrule
% ======= PINN =======
\textbf{PINN} & Mean violation & 0.0904 & 0.0552 & 0.0313 & 0.0168 & 0.0171 & 0.014 & \textbf{0.0164}  & \textbf{0.013} \\
& MSE                 & 0.0104 & 0.0169 & 0.000303 & 0.000505 & 0.000169 & 0.000343 & \textbf{0.000142}  & \textbf{0.000285} \\
& Coverage Width      & 0.694 & 0.489 & \textbf{0.085}  & \textbf{0.0499} & 0.0627 & 0.0432 & \textbf{0.0602}  & \textbf{0.0384} \\
& Coverage Ratio      & 1.00 & 0 & 0.895 & 0.303 & \textbf{0.893}  & \textbf{0.296} & \textbf{0.878}  & \textbf{0.32} \\
& CRPS                & 0.0581 & 0.044 & 0.00948 & 0.00644 & 0.00647 & 0.00579 & \textbf{0.00606}  & \textbf{0.00523} \\
\bottomrule
\end{tabular*}
\end{table*}

\section{Conclusions}

By leveraging the conjugacy of linear Gaussian systems, we derived a closed-form expression to project the output of a probabilistic neural network. This enables the model to remain both expressive and tractable, while embedding domain knowledge directly into the model output.

Empirically, we demonstrated that our framework yields superior performance when data availability is limited. It improves predictive accuracy and uncertainty, incorporating prior knowledge when it is most valuable. When the volume of data increases, the performance gap between constrained and unconstrained models diminishes, although the conditioning step still enables faster training and inference. As a result, further work could investigate utilizing the conditioning step, as a `warm start' to training large neural networks on systems exhibiting linear equality relationships. This is since the conditioning step quickly converges to good values of the parameters, with the unconstrained model then able to fine-tune predictions. As the CPNN also enabled faster inference when compared to their unconstrained counterparts, this provides a promising direction for use within real-time control and optimization on larger scale systems. %Faster training frameworks provides an encouraging avenue for use within these problem settings.

While this study focused on linear equality constraints due to their analytical tractability, many real-world problems indeed involve nonlinear and inequality constraints. Extending the proposed framework to these more generalized constraints, potentially through iterative solvers or surrogate constraint approximations, is a natural avenue for future work. As one of the main advantages of this approach is the reduced computational overhead, the iterative methods may reduce this effect. Moreover, while our framework assumed Gaussian predictive distributions, future extensions could explore non-Gaussian likelihoods, such as heavy-tailed or multi-modal distributions, provided tractable constraint conditioning can still be achieved.

Another promising direction lies in integrating this approach with dynamic systems modelling. For instance, similar approaches could investigate embedding the conditioning framework within more sequential models, such as transformers or recurrent neural networks, to enable more complex (multistep) prediction. %Other data-driven modeling approaches, such as Neural ODEs, could benefit from constraint-aware probabilistic models that enforce known constraints within gradients themselves. 
This opens up applications in areas such as robotics, healthcare, and scientific machine learning, where structure, interpretability, and uncertainty quantification are all critical.

\small
\begin{ack}
MM gratefully acknowledges funding from EPSRC studentship EP/W524323/1. BC gratefully acknowledges funding by EPSRC under grant EP/W003317/1.
\end{ack}

% \section*{DECLARATION OF GENERATIVE AI AND AI-ASSISTED TECHNOLOGIES IN THE WRITING PROCESS}
% During the preparation of this work the author(s) used [NAME TOOL / SERVICE] in order to [REASON]. After using this tool/service, the author(s) reviewed and edited the content as needed and take(s) full responsibility for the content of the publication.

\small
\bibliography{IFAC2026}             % bib file to produce the bibliography

@inproceedings{optuna_2019,
	title={Optuna: A Next-generation Hyperparameter Optimization Framework},
	author={Akiba, Takuya and Sano, Shotaro and Yanase, Toshihiko and Ohta, Takeru and Koyama, Masanori},
	booktitle={Proceedings of the 25th {ACM} {SIGKDD} International Conference on Knowledge Discovery and Data Mining},
	year={2019}
}

@article{pytorch_2019,
title = {{PyTorch}: An Imperative Style, High-Performance Deep Learning Library},
author = {Paszke, Adam and Gross, Sam and Massa, Francisco and Lerer, Adam and Bradbury, James and Chanan, Gregory and Killeen, Trevor and Lin, Zeming and Gimelshein, Natalia and Antiga, Luca and Desmaison, Alban and Kopf, Andreas and Yang, Edward and DeVito, Zachary and Raison, Martin and Tejani, Alykhan and Chilamkurthy, Sasank and Steiner, Benoit and Fang, Lu and Bai, Junjie and Chintala, Soumith}, 
  journal = {Advances in Neural Information Processing Systems},
  volume = {32},
pages = {8024--8035},
year = {2019},
optpublisher = {Curran Associates, Inc.},
opturl = {http://papers.neurips.cc/paper/9015-pytorch-an-imperative-style-high-performance-deep-learning-library.pdf}
}

@misc{marshLearningEmbeddedLinear2026a,
  title = {Learning with {{Embedded Linear Equality Constraints}} via {{Variational Bayesian Inference}}},
  author = {Marsh, Matthew and Chachuat, Beno{\^i}t and Chanona, Antonio del Rio},
  year = 2026,
  publisher = {arXiv},
  doi = {10.48550/ARXIV.2604.24911},
  urldate = {2026-05-14},
  copyright = {Creative Commons Attribution 4.0 International},
  langid = {english}
}

@incollection{agrawalDifferentiableConvexOptimization2019,
  title = {Differentiable Convex Optimization Layers},
  booktitle = {Proceedings of the 33rd {{International Conference}} on {{Neural Information Processing Systems}}},
  author = {Agrawal, Akshay and Amos, Brandon and Barratt, Shane and Boyd, Stephen and Diamond, Steven and Kolter, J. Zico},
  year = 2019,
  month = dec,
  number = {858},
  pages = {9562--9574},
  optpublisher = {Curran Associates Inc.},
  optaddress = {Red Hook, NY, USA},
  urldate = {2025-04-25}
}

@article{ahmedComparativeStudyMachine2025,
  title = {Comparative {{Study}} of {{Machine Learning}} and {{System Identification}} for {{Process Systems Engineering Dynamics}}},
  author = {Ahmed, Akhil and {del Rio-Chanona}, Ehecatl Antonio and Mercang{\"o}z, Mehmet},
  year = 2025,
  month = feb,
  journal = {Industrial \& Engineering Chemistry Research},
  volume = {64},
  number = {8},
  pages = {4450--4478},
  publisher = {American Chemical Society},
  issn = {0888-5885},
  urldate = {2025-04-08}
}

@misc{balestrieroPOLICEProvablyOptimal2023,
  title = {{{POLICE}}: {{Provably Optimal Linear Constraint Enforcement}} for {{Deep Neural Networks}}},
  shorttitle = {{{POLICE}}},
  author = {Balestriero, Randall and LeCun, Yann},
  year = 2023,
  month = mar,
  note = {arXiv:2211.01340},
  eprint = {2211.01340},
  primaryclass = {cs},
  publisher = {arXiv},
  urldate = {2025-04-25},
  archiveprefix = {arXiv}
}

@article{chenPhysicsinformedNeuralNetworks2024,
  title = {Physics-Informed Neural Networks with Hard Linear Equality Constraints},
  author = {Chen, Hao and Flores, Gonzalo E. Constante and Li, Can},
  year = 2024,
  month = oct,
  journal = {Computers \& Chemical Engineering},
  volume = {189},
  pages = {108764},
  issn = {0098-1354},
  urldate = {2025-04-25}
}

@misc{dontiDC3LearningMethod2021,
  title = {{{DC3}}: {{A}} Learning Method for Optimization with Hard Constraints},
  shorttitle = {{{DC3}}},
  author = {Donti, Priya L. and Rolnick, David and Kolter, J. Zico},
  year = 2021,
  month = apr,
  note = {arXiv:2104.12225},
  eprint = {2104.12225},
  primaryclass = {cs},
  publisher = {arXiv},
  urldate = {2025-04-08},
  archiveprefix = {arXiv}
}

@misc{gawlikowskiSurveyUncertaintyDeep2022,
  title = {A {{Survey}} of {{Uncertainty}} in {{Deep Neural Networks}}},
  author = {Gawlikowski, Jakob and Tassi, Cedrique Rovile Njieutcheu and Ali, Mohsin and Lee, Jongseok and Humt, Matthias and Feng, Jianxiang and Kruspe, Anna and Triebel, Rudolph and Jung, Peter and Roscher, Ribana and Shahzad, Muhammad and Yang, Wen and Bamler, Richard and Zhu, Xiao Xiang},
  year = 2022,
  month = jan,
  note = {arXiv:2107.03342},
  eprint = {2107.03342},
  primaryclass = {cs},
  publisher = {arXiv},
  urldate = {2025-04-08},
  archiveprefix = {arXiv}
}

@misc{gonzalezNeuralNetworksFast2024,
  title = {Neural {{Networks}} for {{Fast Optimisation}} in {{Model Predictive Control}}: {{A Review}}},
  shorttitle = {Neural {{Networks}} for {{Fast Optimisation}} in {{Model Predictive Control}}},
  author = {Gonzalez, Camilo and Asadi, Houshyar and Kooijman, Lars and Lim, Chee Peng},
  year = 2024,
  month = dec,
  note = {arXiv:2309.02668},
  eprint = {2309.02668},
  primaryclass = {eess},
  publisher = {arXiv},
  urldate = {2025-06-16},
  archiveprefix = {arXiv}
}

@inproceedings{hansenLearningPhysicalModels2023,
  title = {Learning {{Physical Models}} That {{Can Respect Conservation Laws}}},
  booktitle = {Proceedings of the 40th {{International Conference}} on {{Machine Learning}}},
  author = {Hansen, Derek and Maddix, Danielle C. and Alizadeh, Shima and Gupta, Gaurav and Mahoney, Michael W.},
  year = 2023,
  month = jul,
  pages = {12469--12510},
  optpublisher = {PMLR},
  issn = {2640-3498},
  urldate = {2025-07-23},
  langid = {english}
}

@misc{iftakherPhysicsInformedNeuralNetworks2025,
  title = {Physics-{{Informed Neural Networks}} with {{Hard Nonlinear Equality}} and {{Inequality Constraints}}},
  author = {Iftakher, Ashfaq and Golder, Rahul and Hasan, M. M. Faruque},
  year = 2025,
  month = jul,
  note = {arXiv:2507.08124},
  eprint = {2507.08124},
  primaryclass = {cs},
  publisher = {arXiv},
  urldate = {2025-07-23},
  archiveprefix = {arXiv}
}

@article{immerEffectiveBayesianHeteroscedastic2023,
  title = {Effective {{Bayesian Heteroscedastic Regression}} with {{Deep Neural Networks}}},
  author = {Immer, Alexander and Palumbo, Emanuele and Marx, Alexander and Vogt, Julia},
  year = 2023,
  month = dec,
  journal = {Advances in Neural Information Processing Systems},
  volume = {36},
  pages = {53996--54019},
  urldate = {2025-10-03},
  langid = {english}
}

@misc{lastrucciENFORCEExactNonlinear2025,
  title = {{{ENFORCE}}: {{Exact Nonlinear Constrained Learning}} with {{Adaptive-depth Neural Projection}}},
  shorttitle = {{{ENFORCE}}},
  author = {Lastrucci, Giacomo and Schweidtmann, Artur M.},
  year = 2025,
  month = feb,
  note = {arXiv:2502.06774},
  eprint = {2502.06774},
  primaryclass = {cs},
  publisher = {arXiv},
  urldate = {2025-04-08},
  archiveprefix = {arXiv}
}

@article{lawrenceMachineLearningIndustrial2024,
  title = {Machine Learning for Industrial Sensing and Control: {{A}} Survey and Practical Perspective},
  shorttitle = {Machine Learning for Industrial Sensing and Control},
  author = {Lawrence, Nathan P. and Damarla, Seshu Kumar and Kim, Jong Woo and Tulsyan, Aditya and Amjad, Faraz and Wang, Kai and Chachuat, Benoit and Lee, Jong Min and Huang, Biao and Bhushan Gopaluni, R.},
  year = 2024,
  month = apr,
  journal = {Control Engineering Practice},
  volume = {145},
  pages = {105841},
  issn = {0967-0661},
  urldate = {2025-11-18}
}

@article{raissiPhysicsinformedNeuralNetworks2019,
  title = {Physics-Informed Neural Networks: {{A}} Deep Learning Framework for Solving Forward and Inverse Problems Involving Nonlinear Partial Differential Equations},
  shorttitle = {Physics-Informed Neural Networks},
  author = {Raissi, M. and Perdikaris, P. and Karniadakis, G. E.},
  year = 2019,
  month = feb,
  journal = {Journal of Computational Physics},
  volume = {378},
  pages = {686--707},
  issn = {0021-9991},
  urldate = {2025-04-16}
}

@misc{runjeConstrainedMonotonicNeural2023,
  title = {Constrained {{Monotonic Neural Networks}}},
  author = {Runje, Davor and Shankaranarayana, Sharath M.},
  year = 2023,
  month = may,
  note = {arXiv:2205.11775},
  eprint = {2205.11775},
  primaryclass = {cs},
  publisher = {arXiv},
  urldate = {2025-04-08},
  archiveprefix = {arXiv}
}

@misc{seitzerPitfallsHeteroscedasticUncertainty2022,
  title = {On the {{Pitfalls}} of {{Heteroscedastic Uncertainty Estimation}} with {{Probabilistic Neural Networks}}},
  author = {Seitzer, Maximilian and Tavakoli, Arash and Antic, Dimitrije and Martius, Georg},
  year = 2022,
  month = apr,
  note = {arXiv:2203.09168},
  eprint = {2203.09168},
  primaryclass = {cs},
  publisher = {arXiv},
  urldate = {2025-05-08},
  archiveprefix = {arXiv}
}

@misc{tordesillasRAYENImpositionHard2023,
  title = {{{RAYEN}}: {{Imposition}} of {{Hard Convex Constraints}} on {{Neural Networks}}},
  shorttitle = {{{RAYEN}}},
  author = {Tordesillas, Jesus and How, Jonathan P. and Hutter, Marco},
  year = 2023,
  month = jul,
  note = {arXiv:2307.08336},
  eprint = {2307.08336},
  primaryclass = {cs},
  publisher = {arXiv},
  urldate = {2025-04-08},
  archiveprefix = {arXiv}
}

@misc{utkarshEndtoEndProbabilisticFramework2025,
  title = {End-to-{{End Probabilistic Framework}} for {{Learning}} with {{Hard Constraints}}},
  author = {Utkarsh, Utkarsh and Maddix, Danielle C. and Ma, Ruijun and Mahoney, Michael W. and Wang, Yuyang},
  year = 2025,
  month = jun,
  note = {arXiv:2506.07003},
  eprint = {2506.07003},
  primaryclass = {cs},
  publisher = {arXiv},
  urldate = {2025-07-23},
  archiveprefix = {arXiv}
}

@article{villanuevaStabilitySetValuedIntegration2014,
  title = {On the {{Stability}} of {{Set-Valued Integration}} for {{Parametric Nonlinear ODEs}}},
  journal = {Computer Aided Chemical Engineering},
  author = {Villanueva, Mario E. and Houska, Boris and Chachuat, Beno{\^i}t},
  editor = {Kleme{\v s}, Ji{\v r}{\'i} Jarom{\'i}r and Varbanov, Petar Sabev and Liew, Peng Yen},
  year = 2014,
  month = jan,
  series = {24 {{European Symposium}} on {{Computer Aided Process Engineering}}},
  volume = {33},
  pages = {595--600},
  publisher = {Elsevier},
  urldate = {2025-08-22}
}
                                                     % with bibtex (preferred)

% \begin{table}[h!]
% \begin{center}
% \caption{Reactor variables and parameters}\label{tb:reactor_variables}
% \label{batch_params}
% \begin{tabular}{lll}
% \toprule
% \textbf{Symbol} & \textbf{Description} & \textbf{Units} \\
% \midrule
% $C_A, C_B, C_C$ & Concentrations of A, B, C & kmol/m³ \\
% $T, T_c$        & Reactor, coolant temperature & K \\
% $r_1, r_2$      & Reaction rates & kmol/(m³·s) \\
% $k_{0i}, E_{Ai}$ & Kinetic parameters & 1/s, kJ/kmol \\
% $\Delta H_i$    & Reaction enthalpies & kJ/kmol \\
% $\rho, C_p$     & Density, heat capacity & kg/m³, kJ/(kg·K) \\
% $UA, V$         & Heat transfer, volume & kJ/(s·K), m³ \\
% \bottomrule
% \end{tabular}
% \end{center}
% \end{table}

\end{document}